\pdfoutput=1

\documentclass[11pt]{article}

\usepackage[]{ACL2023}

\usepackage{times}
\usepackage{latexsym}
\usepackage[utf8]{inputenc}
\usepackage{microtype}
\usepackage{inconsolata}
\usepackage{graphicx} 
\usepackage{booktabs}
\usepackage{tablefootnote}
\usepackage{enumitem}
\usepackage{amssymb}
\usepackage{adjustbox}
\usepackage{multirow}
\usepackage{CJKutf8} 
\usepackage{blindtext} 
\usepackage{array} 

\newcommand{\method}{Panel of LLm evaluators\ }
\newcommand{\methodcaps}{Panel of LLM Evaluators\ }
\newcommand{\acronymnospace}{PoLL}
\newcommand{\acronym}{PoLL\ }

\makeatletter
\def\blfootnote{\xdef\@thefnmark{}\@footnotetext}
\makeatother
\title{Replacing Judges with Juries: \\Evaluating LLM Generations with a Panel of Diverse Models}

\author{Pat Verga \\ {\bf Sebastian Hofstätter, Sophia Althammer, Yixuan Su} \\ {\bf Aleksandra Piktus, Arkady Arkhangorodsky, Minjie Xu, Naomi White} \\{\bf Patrick Lewis} \\
        Cohere \\ 
        }

\begin{document}

\maketitle

\begin{abstract}
\end{abstract}
As Large Language Models (LLMs) have become more advanced, they have outpaced our abilities to accurately evaluate their quality. Not only is finding data to adequately probe particular model properties difficult, but evaluating the correctness of a model's free-form generation alone is a challenge. To address this, many evaluations now rely on using LLMs themselves as judges to score the quality of outputs from other LLMs. Evaluations most commonly use a single large model like GPT-4. While this method has grown in popularity, it is costly, has been shown to introduce intra-model bias, and in this work, we find that very large models are often unnecessary. We propose instead to evaluate models using a \method (\acronymnospace). Across three distinct judge settings and spanning six different datasets, we find that using a \acronym composed of a larger number of smaller models outperforms a single large judge, exhibits less intra-model bias due to its composition of disjoint model families, and does so while being over seven times less expensive. 

\section{Introduction}
Evaluating generative language models is a challenging task: not only is it difficult to find meaningful data to test the models, but evaluating the correctness of a generated response is itself a challenge. Multiple choice datasets like MMLU~\cite{hendrycks2020measuring} have become popular in part by side-stepping the difficulty of evaluating generations. However, multiple-choice questions are in many ways probing a different property than that of a free-form generative task, which is oftentimes closer to the downstream use-case. 

\begin{figure}[ht]
    \centering
    \includegraphics[width=0.5\textwidth]{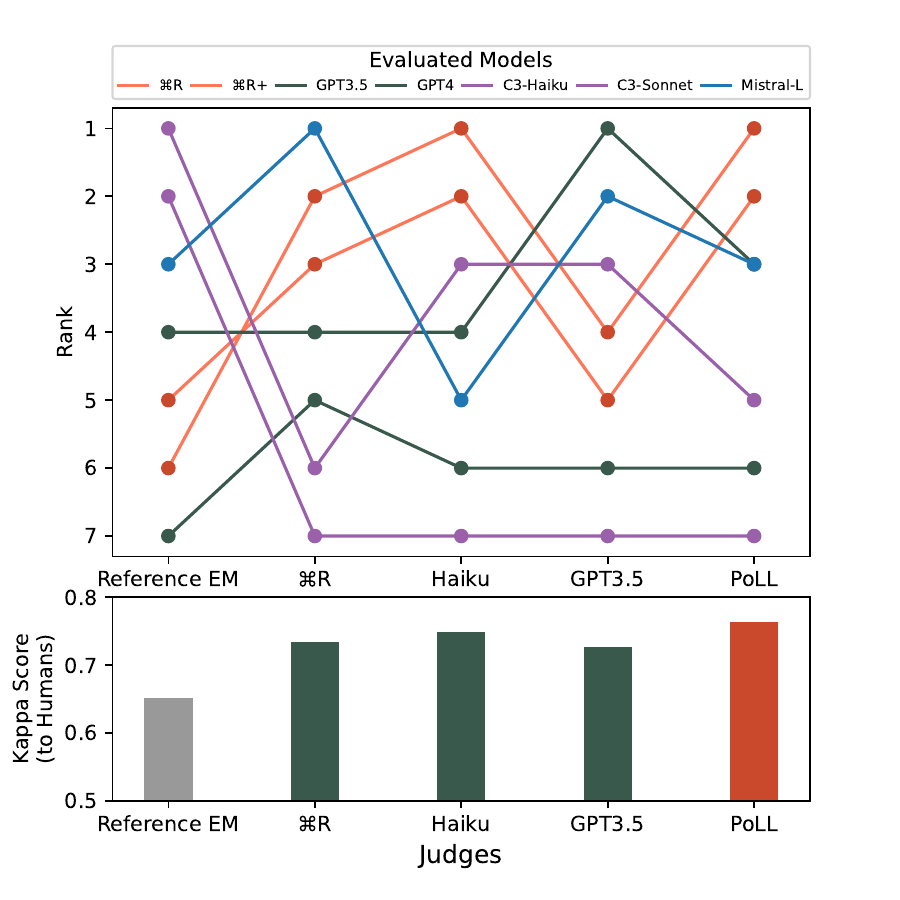}
    \caption{Top: Rankings of model performance change drastically depending on which LLM is used as the judge on KILT-NQ. Bottom: The \method (\acronymnospace) has the highest Cohen's $\kappa$ correlation with human judgements.}
    \label{fig:nq_ranking}
\end{figure}

Many automatic metrics have been used across various tasks such as BLEU in machine translation~\citep{papineni2002bleu}, ROUGE for summarization~\citep{lin-2004-rouge}, and heuristic string match methods, such as normalized exact match (EM) and token level F1 for question answering~\cite{rajpurkar2016squad}. However, these simplistic methods commonly fail to analyze the intended property of interest. QA metrics, for example, invariably lead to both false positive failures (e.g. superfluous token overlap) and more commonly false negatives due to an incomplete set of gold reference answers (e.g. date format differences\footnote{We found that EM unjustly penalized Command models for a tendency to write in Canadian or British English as QA dataset annotations typically format dates in American MM-DD-YYYY format.}, inclusion of middle initial in person's name, etc.).

More recent methods have attempted to address these issues by instead using trained or prompted models as evaluators~\cite{sellam-etal-2020-bleurt, zheng2024judging, li2024leveraging, kocmi-federmann-2023-large, shen-etal-2023-large, li2023prd}. Prior work has shown that model-based scoring methods often correlate better with human judgements than heuristic metrics like EM~\cite{bohnet2022attributed,zheng2024judging} and that strong evaluator models generalize well across different tasks~\cite{huang2024empirical}. 

Unfortunately, while the use of LLMs like GPT-4 as evaluators has become increasingly common, it has also been observed that evaluator models tend to have their own biases; often recognizing and preferring their own outputs over those of other models~\cite{Panickssery2024LLMEval}. Additionally, it is most common to use the largest, most universally capable models as evaluators, which is both slow and costly, limiting applicability and access. 

In this paper, we perform experiments across three settings (single-hop QA, multi-hop QA, and Chatbot Arena), spanning six datasets, and make the following contributions:
\begin{enumerate}
\item We propose to evaluate LLM generations using a \method (\acronymnospace) drawn from different model families rather than a single large judge (Section \ref{sec:method}).
\item We show that using an instantiation of \acronym correlates better with human judgements compared to a single large judge (GPT-4), while being over seven times cheaper (Sections \ref{sec:results_kappa} and \ref{sec:result_arena}).
\item In some scenarios, GPT-4 is a relatively weak judge, exhibiting high variance with minor changes to the prompt (Section \ref{sec:prompt_variance}).
\item Intra-model scoring bias is reduced by pooling judgements across a panel of heterogeneous evaluator models (Section \ref{sec:result_bias}).
\end{enumerate}

\section{Methods \label{sec:method}}
\subsection{Background: LLM as a Judge \label{sec:judge_background}}
A judge evaluator model $J$ is used to score the output $a$ from a test model $A$. 

\noindent\textbf{Single-point Scoring}
In some settings, $J$ is tasked with rating the quality of a single model output independently of any point of comparison~(e.g.~\citep{kocmi2023large}). $J_{\textup{prompt}}$ will often include natural language instructions on how the grading should be performed (i.e. what properties constitute a good or bad output). Aside from the prompt instructions, the rating is based solely on $J$'s internal model of what a quality output is. Here, $\textup{score}=J(a)$.

\noindent\textbf{Reference-based Scoring}
In other cases, the model is provided with some 'gold' reference $r$, which contains the information that should be included in $a$~(e.g.~\cite{zhu2023judgelm}). For example, in QA the reference would be the 'correct' answer to the question. In this case, $\textup{score}=J(a,r)$. This setting is explored in Sections \ref{sec:single_qa} and \ref{sec:mh_qa}.

\noindent\textbf{Pair-wise Scoring}
Another very common setting is in pair-wise scoring where the goal is to choose which of two outputs is better~(e.g.~\cite{zheng2024judging}). Given outputs $a$ and $b$ generated by two models $A$ and $B$, an evaluator $J$ compares them and generates a preference score over the outputs as $\textup{score}=J(a,b)$\footnote{To avoid ordering bias, it is common to test both $a-b$ and $b-a$ orderings as done in Section \ref{sec:arena_hard}}. The form of the score can vary based on the use-case, but it is often a three or five point scale such as $a>b$, $a\approx b$, $a<b$. This setting is used in Section \ref{sec:arena_hard}.

\subsection{\methodcaps}
The above settings assume that all scoring is performed by a single capable judge. However, as outlined earlier, one of the largest issues with relying on a single model $J$, such as GPT-4, is that it introduces intra-model bias. To address this we instead propose to score answer correctness based not on a single judge, but instead on a panel composed of multiple evaluator models. Similar pooling techniques are used to reduce variance in human annotations by normalizing out both natural variation in human judgements caused by their own subjective biases as well as human error~\cite{vorhees1998variation}. In prior relevant work, \citet{li2023prd} proposed an evaluation method with multiple judges to reduce bias though only looked at large models on pair-wise evaluations.

To calculate the \acronym score, each evaluator model independently scores a given model output just as they would in any of the scenarios outlined above. Those individual scores are then pooled together through a voting function\footnote{In this work we consider both max and average pool voting, though other functions could also work well.} such that the final score  $= f(j \in P: j(a))$ where $P$ is a panel composed of individual judges $j$ and $f$ is a voting function. 

\section{Experimental Settings}

\subsection{\acronym Composition and Voting}
In our experiments, We construct a \acronym from three models being drawn from three disparate model families (Command 
R, Haiku, and GPT-3.5). We consider two different voting functions for aggregating scores across the judges. For QA datasets, we use max voting, as all judgements are binary [correct, incorrect]. For Chatbot Arena we instead use average pooling because judgements are scores ranging from 1-5 and a three judge panel often does not produce a clear majority decision\footnote{We also experimented with max voting with average voting fallback which yielded similar results with an identical overall ranking.}. 

\subsection{Model Families}
\noindent\textbf{Command R Family:}
Command R (CMD-R, 35B) and Command R+ (CMD-R+, 103B) are open weight models created by Cohere\footnote{Release blog of Command R \url{https://cohere.com/blog/command-r} and Command R+ \url{https://cohere.com/blog/command-r-plus-microsoft-azure}}.
We consider Command R as one of the models in the \acronymnospace.

\noindent\textbf{GPT Family:}
GPT-3.5 and GPT-4~\cite{openai2024gpt4} are two of the most widely used models. GPT-4 is typically the default choice for models as judges. We use GPT-3.5 as a member of our \acronymnospace.

\noindent\textbf{Claude-3 Family:}
Models built by Anthropic include (in increasing size) Haiku, Sonnet, and Opus~\cite{anthropic2024sonnet}. We include Haiku in our \acronymnospace. 

\noindent\textbf{Mistral Family:}
Mistral Large (Mistral-LG) and Mistral Medium (Mistral-MD) are proprietary models created by Mistral\footnote{Release blog of Mistral Large: \url{https://mistral.ai/news/mistral-large/} and Mistral Medium: \url{https://mistral.ai/news/la-plateforme/}.}. We did not conduct experiments with Mistral judges but evaluate their generations as a point of comparison to have a model 'unaffiliated' with any judges.

\subsection{Single-hop Question Answering \label{sec:single_qa}}
Our question answering (QA) tasks are open-book settings where a model $m$ is given a question $q$ and must retrieve evidence $e$ from some retrieval system (such as the internet or dense index over wikipedia) and must generate an answer $g$ as $g=m(q, e)$. Each instance is accompanied by a human annotated reference answer. 

We run experiments on datasets from KILT~\cite{petroni2021kilt} versions of Natural Questions (NQ)~\cite{kwiatkowski-etal-2019-natural}, TriviaQA (TQA)~\cite{joshi-etal-2017-triviaqa}, and HotpotQA (HPQA)~\cite{yang-etal-2018-hotpotqa} (see Section \ref{sec:appendix_qa} for more details).

\subsection{Multi-hop Question Answering \label{sec:mh_qa}}
Multi-hop question answering is similar to the single-hop setting described in Section \ref{sec:single_qa}. However, the questions are designed such that models must perform multiple rounds of retrieval to answer sub-questions and collect sufficient evidence to ultimately answer the initial question. We perform experiments on two datasets: Bamboogle~\citep{press2023measuring} and HPQA. See Section \ref{sec:appendex_react} for additional details.

\subsection{Chatbot Arena Hard \label{sec:arena_hard}}
Chatbot Arena\footnote{\url{https://chat.lmsys.org/}} is a popular benchmark for evaluating LLM head-to-head performance. This crowd-sourced effort has users prompt a pair of anonymized LLMs and rate which output is better~\cite{zheng2024judging}. 

Recently,~\citet{arenahard2024} mined a 'hard' subset of 500 arena prompts and created an automatic evaluation where an LLM judge (GPT-4) would score a model output against a baseline generation (coming from a different version of GPT-4) in a head-to-head comparison. In their analysis, they determined that this evaluation correlated better with the human judgements from the original Chatbot Arena annotations when compared against other commonly used automated evaluation benchmarks. 

At the time of writing, the repository only contained generated answers from GPT-3.5 and GPT-4. We therefore generated new outputs for all remaining models and subsequently scored generated answers with the various judges using the original codebase from~\citet{arenahard2024}\footnote{\url{https://github.com/lm-sys/arena-hard}}. The raw scores can be found in Table \ref{tab:arena_hard_scores}. Following the original work, we treat Chatbot Arena crowdsourced annotations as ground truth for calculating correlation between evaluator models and human judgements.

\subsection{Prompting Judges}
As alluded to in Section \ref{sec:judge_background}, judge models need to be prompted in different ways depending on the particular task setup. Our QA experiments use reference-based scoring and our models prompts contain few-shot in-context examples of valid and invalid $q, a, r$ triples. We also experiment with variations on these prompts in Section \ref{sec:prompt_variance}. Further details on judge prompts may be found in the appendix.

As an additional point of reference in our QA experiments, we include the commonly used exact match (EM) metric judgements. We specifically use the 'containment' version of EM from prior work which is more amenable to LLM long-form generation~\citep{liu2024lost} and checks if a reference answer string appears within the generated model response (after normalization).

In our experiments on Chatbot Arena Hard, we follow the same pair-wise scoring setup of the original work. We do not modify the judge prompts in any way from the original implementation.

\subsection{Human Judgements}
To gather human reference judgements, we utilized Cohere's internal highly-qualified annotation workforce. Annotators were shown a single anonymized model generated answer at a time along with the original question and reference answer. Annotators were asked to judge whether the reference answer is semantically contained inside the generated answer\footnote{By asking for answer correctness alone, we hope to minimize any potential annotator bias or preference towards a particular model's style.}.  See Section \ref{sec:appendix_annotation} for additional details 

\section{Results \label{sec:results}}

\subsection{Correlation to Human Judgements \label{sec:results_kappa}}
\noindent\textbf{Cohen's $\kappa$ Correlation}
Cohen's kappa measures inter-rater reliability, which quantifies the level of agreement between two or more raters or judges. The kappa statistic takes into account the possibility of agreement occurring by chance, making it a more robust measure than simple percent agreement.

The formula for Cohen's kappa is:
$$
\kappa = \frac{p_o - p_e}{1 - p_e}
$$

\noindent where $p_o$ is the observed agreement proportion and $p_e$ is the expected agreement proportion by chance. This results in a value of $\kappa$ ranging from -1 to 1. Opinions vary on how scores should be interpreted, but in general $\kappa > 0.8$ is considered a strong correlation and $\kappa > 0.6$ is a moderate correlation.

In Table \ref{tab:judge_model_performance} we can see how the ratings from different evaluator judges, on different single-hop QA datasets from KILT, correlate with human judgements as measured by $\kappa$. We see that overall, \acronym has the strongest correlation across various tasks, while GPT-4 is one of the weaker evaluators on this particular task setup (See \ref{sec:prompt_variance} for further analysis).

\begin{table}[ht]
\centering
\begin{tabular}{lccc}
\toprule
\textbf{Judge} & \textbf{NQ} & \textbf{TQA} & \textbf{HPQA} \\ \midrule
EM & 0.651 & 0.827 & 0.662 \\ 
GPT-4 & 0.627 & 0.841 & 0.830 \\ 
\midrule
CMD-R & 0.734 & \underline{0.902} & 0.815 \\ 
Haiku & \underline{0.749} & 0.894 & \textbf{0.873} \\ 
GPT-3.5 & 0.726 & 0.859 & 0.833 \\ 
\midrule
\acronym & \textbf{0.763} & \textbf{0.906} & \underline{0.867} \\
\bottomrule
\end{tabular}
\caption{Cohen's Kappa Judge Model Performance on Different Single-hop QA Datasets from KILT, where best results are indicated by the bold font and second best results are underlined.}
\label{tab:judge_model_performance}
\end{table}

\begin{table}[ht]
\centering
\begin{tabular}{lcc}
\toprule
\textbf{Judge/Correlation} & Pearson & Kendall Tau \\ 
\midrule
GPT-4&0.817&0.667\\
\midrule
Haiku&0.883&0.722\\
GPT-3.5&0.883&0.730\\
CMD-R&0.817&0.676\\
\midrule
\acronym &\textbf{0.917}&\textbf{0.778}\\
\bottomrule
\end{tabular}
\caption{Pearson and Kendall-Tau correlations between different judge models as compared to the rankings produced by the Chatbot Arena overall leaderboard.}
\label{table:model_correlation}
\end{table}

\subsection{Rank Correlation on Chatbot Arena \label{sec:result_arena}}
\begin{figure}
    \centering
    \includegraphics[width=0.48\textwidth]{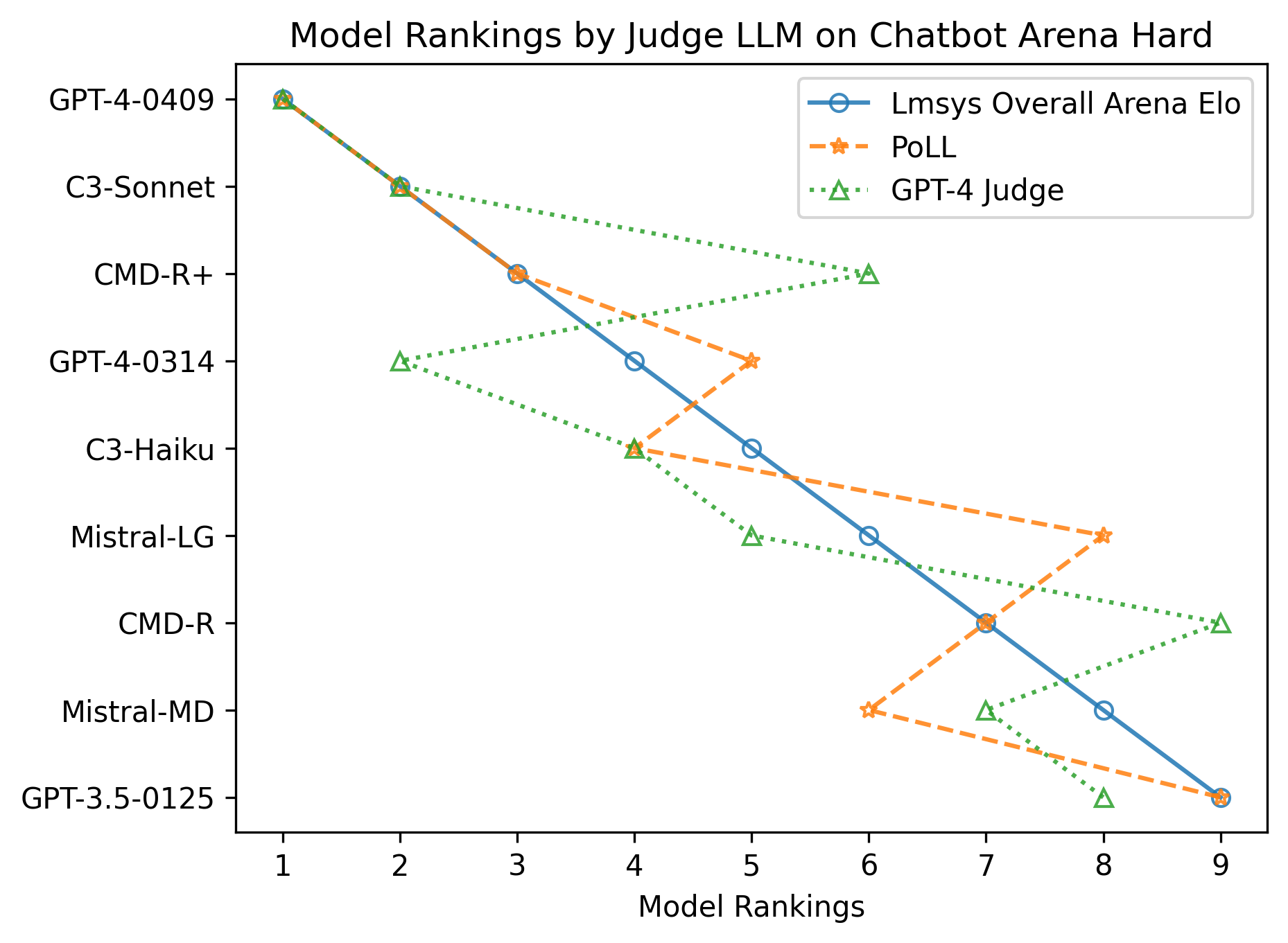}
    \caption{Rankings of model performance on Chatbot Arena Hard judged by GPT-4 or \acronymnospace. Ranks are compared to those in the original Chatbot Arena.}
    \label{fig:arena_hard_ranks}
\end{figure}

In Table \ref{table:model_correlation} we can see how different judge model rankings correlate with human judgement. Following~\citet{arenahard2024}, we treat the crowd-sourced ELO rankings from Chatbot Arena as ground truth. We calculate both Kendall Tau~\citep{kendall1938new} and Pearson Correlation~\citep{pearson1895vii} of the ranked list produced by each of the judge methods with respect to this ground truth ranking.  We find that \acronym is best correlated with the gold rankings, particularly at the top of the ranked list as shown in Figure \ref{fig:arena_hard_ranks}. 

\subsection{Judgement Variance by Prompt Changes \label{sec:prompt_variance}}
Based on the observation that GPT-4 was the weakest judge model on our KILT evaluations, we investigated how the model reacts to modifications to its prompt. GPT-4 is the most powerful judge model we tested, yet it performed worse than less capable models on what is essentially a fuzzy string matching exercise. We hypothesize that may be because GPT-4 is over-reasoning and injecting too much background knowledge into determining the correctness of an answer rather than simply aligning the gold reference with the generation. In Table \ref{tab:prompt_variants}, we can see how the correlation between GPT-4 and human annotators varies as the prompt changes. In all cases, having in-context examples improves the performance over zero-shot and the most effective strategy is an explicit instruction to the model not to 'overthink' and not to concern itself with the wider factuality of the answers with respect to the outside world.\footnote{the 'don't overthink' prompt had a negligible positive effect on Haiku, minor negative effect to Command R and a severe negative impact on GPT-3.5.} These changes bring the agreement level for GPT-4 up to the level of GPT-3.5 when using our few-shot standard prompt, though still below Command-R, Haiku, and \acronymnospace.

\begin{table}[ht]
\centering
\begin{tabular}{lc}
\toprule
\textbf{Prompt Variant} & \textbf{Kappa} \\ \midrule
Zero-shot & 0.518 \\
Few-Shot Standard & 0.627 \\
+No Instruction Line & 0.594 \\
+Move Instruction Line & 0.637 \\
+Chat Formatted Shots & 0.561 \\
+'don't overthink'& 0.725 \\ 
\bottomrule
\end{tabular}
\caption{Kappa values on NQ for different prompt variants with GPT-4 as judge. \label{tab:prompt_variants}. For reference, GPT3.5 with  the few-shot standard prompt achieves 0.726 (table \ref{tab:judge_model_performance})}
\end{table}

\subsection{Judge Bias and Consistency \label{sec:result_bias}}
\begin{figure}
    \centering
    \includegraphics[width=0.5\textwidth]{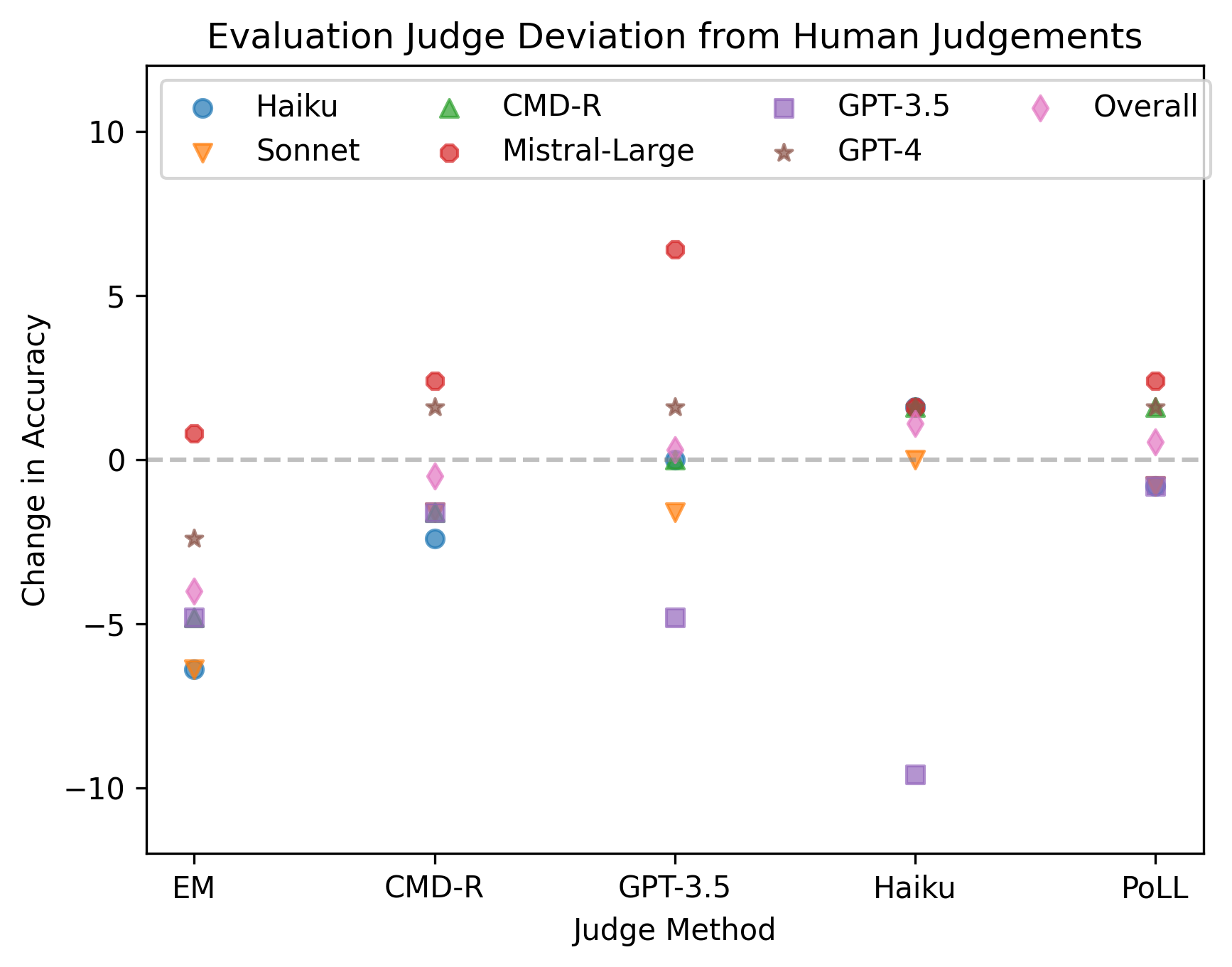}
    \caption{Accuracy changes of different evaluation judges as compared to human judgements on HotPotQA (multi-hop).}
    \label{fig:hpqa_multi_scatter}
\end{figure}

\begin{figure}
    \centering
    \includegraphics[width=0.5\textwidth]{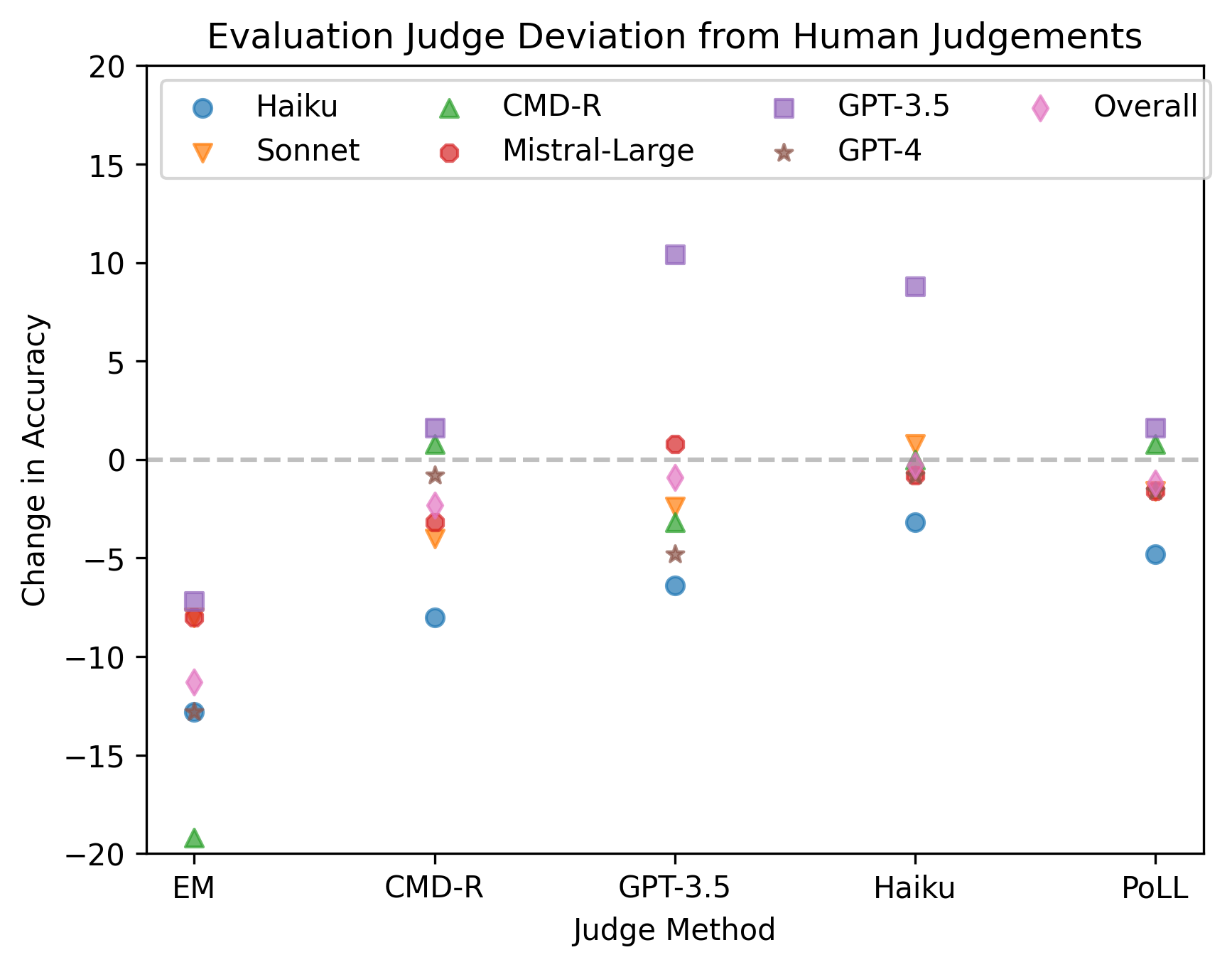}
    \caption{Accuracy changes of different evaluation judges as compared to human judgements on Bamboogle.}
    \label{fig:bamboogle_scatter}
\end{figure}

One of the biggest motivators for replacing a single large judge with a panel of heterogeneous models is to reduce bias in evaluation. To analyze to what extent this was the case, we compared the delta in absolute accuracy score for our individual judges and \acronym relative to scores by human annotators across our multi-hop datasets. Figures \ref{fig:hpqa_multi_scatter} and \ref{fig:bamboogle_scatter} show results on HotPotQA and Bamboogle. We can see how the different judges score different models and how far those predictions deviate from human annotator decisions (the dotted line at 0). We observe that overall, \acronym has the smallest spread in scores, with a standard deviation of 2.2, compared to EM and individual judges. GPT-3.5 has the highest spread, with a standard deviation of 6.1. We also see in Figure \ref{fig:bamboogle_scatter} that the highest positive delta for each individual model being scored occurs when it is judged by itself.

In Figure \ref{fig:arena_hard_ranks}, we assessed the model rankings produced on Chatbot Arena Hard by \acronym as compared to the GPT-4 judge from the original work. The 'gold' ranking appears on the diagonal and represents the rankings from the original Chatbot Arena ELO. We find that \acronym rankings correlate better with the ground truth, particularly at the top of the ranked list. We can clearly observe intra-model bias as the GPT-4 judge ranks another GPT-4 variant in position 2, higher than its actual position 4, which is in line with previous works that have also observed GPT-4's preference for its own generations~\cite{zheng2024judging,Panickssery2024LLMEval}

\subsection{Cost and Latency}
At the time of writing, the cost of running our specific instance of \acronym is \$1.25/input\footnote{per million tokens} + \$4.25/output, whereas the cost of running GPT-4 Turbo is \$10/input + \$30/output. Depending on the ratio of input-to-output tokens in a given task, running the entire three model \acronym is seven to eight times less expensive than running a single GPT-4 judge. 

We did not run explicit latency evaluations and many factors such as model choice, serving platform, and more can impact speed. In general though, running a collection of smaller models in parallel (as in \acronymnospace) is faster than a single big model.




\section{Conclusions and Limitations}
In this paper, we showed how a \methodcaps composed of smaller models is not only an effective method for evaluating LLM performance, but also reduces intra-model bias, latency, and cost. The benefits of \acronym are bolstered by the finding that there is not a single 'best' judge across all settings, while \acronym performs well consistently.

In this work we investigated only three evaluator settings and a limited number of judges and panel compositions. While we showed that \acronym is an effective alternative to a single large model in these settings, further work is needed to see how broadly applicable the method is, for example, in math or reasoning evaluations, where language models often struggle~\cite{zheng2024judging}. We also leave the task of 'panel selection', or identifying the best models to include in \acronym in terms of quality and cost, to future work.

\section*{Acknowledgement}
We'd like to thank all of the team at Cohere and in particular Maxime Voisin, Victoria Spooner, Priyanka Sen, Harry Moynehan, Erik Arakelyan, Antoine Debugne, Alex Barbet, Max Bartolo, Acyr Locatelli, Sara Hooker, and Phil Blunsom. Thanks also to Emma Strubell.

\bibliography{anthology,custom}
\bibliographystyle{acl_natbib}

\appendix
\label{sec:appendix}
\section{Appendix}
\subsection{Additional Experimental Details}
\subsubsection{Single-hop QA \label{sec:appendix_qa}}
\noindent\textbf{Datasets}
KILT~\cite{petroni2021kilt} is a popular benchmark that measures a model's open domain question-answering capabilities with 1) a retrieval step to retrieve relevant documents from a knowledge source and 2) an answer generation step taking into account the relevant documents from (1). We select three popular test sets from the KILT benchmark, Natural Questions (NQ)~\cite{kwiatkowski-etal-2019-natural}, TriviaQA (TQA)~\cite{joshi-etal-2017-triviaqa}, and HotpotQA (HPQA)~\cite{yang-etal-2018-hotpotqa} for evaluation. The Natural Questions test set consists of real user questions sampled from a popular web search engine. 
TriviaQA is a test set with complex questions written by trivia enthusiasts. 
HotpotQA is a multi-hop question answering test set where questions require finding and reasoning over multiple supporting documents to answer. 
For all three test sets, the knowledge source is the corpus of Wikipedia documents form the KILT benchmark~\cite{petroni2021kilt}.

\paragraph{KILT Experimental Details:} For KILT, we use the DPR-formatted 100-word chunked KILT wikipedia dump~\cite{Karpukhin2020DensePR,petroni2021kilt}.\footnote{The pre-chunked dump can be downloaded here: \url{https://github.com/facebookresearch/DPR/issues/186\#issuecomment-923268677}}
Then, for all questions, we retrieve and cache the top-10 snippets from the dump, using the GTE-large dense retrieval model~\citep{Wang2023ImprovingTE}, ready to be used to generate answers from different models.

\paragraph{Generating answers for judgement:} To generate answers from all considered models, we ask the question as an unmodified chat message and pass in the snippets using the dedicated \texttt{documents} parameter of the model family's API where available (Command-R Family)\footnote{The prompt template that gets used when using this parameter is described at \url{https://huggingface.co/CohereForAI/c4ai-command-r-v01} with additional details at \url{https://docs.cohere.com/docs/prompting-command-r}}. 
For model families that do no have a specific \texttt{documents} api parameter (GPT, we instead adopt the question answering prompt template used in~\citet{liu2024lost}, which is shown in table~\ref{tab:litm_prompt}.

\paragraph{Judge Prompts} We use the same prompt-template structure for KILT judgements as for Multihop judgements. We initially observed relatively-lower levels of human-agreement when using the multihop prompt verbatim applied to KILT results experiments. 
We hypothesise this was due to mismatch in model-generated answer styles in the few-shot examples.
We therefore create new judgement prompts the following procedure for each of the KILT datasets, randomly sampling a set of labelled fewshot examples from our pool human-labelled examples according to the following constraints:
\begin{itemize}
    \item at least 1 annotated positive example from each model family's generations
    \item at least 1 annotated negative example from each model family's generations
    \item at least 1 example where the human disagrees with keyword-based exact match score.
\end{itemize}

We sweep over \texttt{n\_shots} in $\{8,10,16\}$, picking the prompt that has strongest agreement on average across all judges for a small held-out portion of human labels. The resuling judgement prompts used for KILT in our experiments in the main paper are listed in tables \ref{tab:hotpotqa_kilt_prompt}, \ref{tab:nq_kilt_prompt} and \ref{tab:triviaqa_kilt_prompt}.

\paragraph{GPT-4 judge prompt ablation} Here we include some additional details on prompt ablation for KILT NaturalQuestions mentioned in section \ref{sec:prompt_variance}:
\begin{itemize}
    \item \textbf{Zero-shot}: the natural language instruction is supplied as system call. We then directly ask the model to judge with no labelled examples
    \item \textbf{Few-shot Standard}: This is the prompt in table \ref{tab:nq_kilt_prompt} which is used for all other model judges.
    \item \textbf{No instruction Line}: Here we remove the natural language instruction from the Few-shot standard prompt, on the hypothesis the instruction is confusing the model. This hypothesis turns out to be false as agreement actually drops. (-0.03 $\Delta \kappa$)
    \item \textbf{Move instruction Line}: Here we modify the standard fewshot prompt by moving the instruction line into a separate system call. This results in a small improvement (+0.01 $\Delta \kappa$)
    \item \textbf{Chat-formatted shots}: Here we modify the standard fewshot prompt by formatting each fewshot example as a conversational turn between the user and the assistant. This reduces performance (-0.07 $\Delta \kappa$)
    \item \textbf{`don't overthink'}: Here, we replace the instruction line from the standard prompt's \texttt{You will be given a Question and a Provided Answer. Judge whether the Provided Answer is correct by comparing it to the Reference Answer. Differently formatted dates, people with missing middle names, and alternative spellings should all be considered the same. If the Provided Answer is correct say exactly "True", otherwise say "False".} to a wording which is intended to encourage the model to perform a simpler function and not incorporate external knowledge: \texttt{ "You are judging whether a model has generated a correct answer to a question. Study the examples the user gives you as they will be very informative for how to do the task. The Reference Answers you get will be short. An model's answer will be longer, and can be considered correct if it contains the semantic content of short reference answer somewhere within it. Don't worry about factuality with respect to the real world, just judge the example based on what you see. No need to overthink this task, it really comes down to just soft matching.} This improves results for GPT-4 by +0.07 $\Delta \kappa$. Additional small surface level changes and moving the instruction to a system call lead to an additional +0.03 $\Delta \kappa$. The final optimized prompt for GPT-4 can be found in table \ref{tab:nq_kilt_prompt_gpt4_opt}.
\end{itemize}

The agreements for different models using the prompt optimized for GPT-4 can be found in table \ref{tab:standard_vs_gpt4_prompt}.

\begin{table}[ht]
\centering

\begin{tabular}{lccccc}
\toprule
 & Standard & 'don't overthink' \\
\midrule
EM & 0.651 &  0.651 \\
GPT4 & 0.627 & 0.725 \\
\midrule
CMD-R & 0.734 & 0.687 \\
Haiku & 0.749 & 0.757\\
GPT-3.5 & 0.729 & 0.509\\
\acronym & \textbf{0.763} & 0.699\\
\bottomrule
\end{tabular}

\caption{Kappa values for KILT NQ for the standard prompt and the prompt optimized for GPT-4 (note that the EM baseline is not LLM-dependent and therefore not dependent on prompt)}
\label{tab:standard_vs_gpt4_prompt}

\end{table}

\subsubsection{Multihop}
\textbf{Bamboogle}~\citep{press2023measuring} is 125 questions adversarially constructed to require multiple internet searches to successfully answer. \textbf{HotpotQA} asks questions meant to be answered with an accompanying wikipedia dump (we consider the subset used in~\citep{shinn2024reflexion}).

\label{sec:appendex_react}
\paragraph{Generating answers for judgement -- React Agents:}
All models follow the basic REACT setup for tool-use agents which incorporate a Thought-Action-Observation loop to iteratively call tools and reason over their outputs to find a final answer~\citep{yao2022react}. We use LangChain tool-use implementations for each model when available. On Bamboogle, agents have access to an internet search tool and on HotPotQA, agents have access to a search tool over wikipedia that does dense embedding search using Cohere EmbedV3 + Rerank 3. All search tool calls return the top two most relevant documents to the model.

\paragraph{Judge Prompt} The prompt used for judging multihop answers can be found in table \ref{tab:multihop_judge_prompt}

\subsection{Additional Results}
\subsubsection{Multi-hop QA}

In Tables \ref{tab:kappa-mh-bamboogle} and \ref{tab:kappa-mh-hpqa} we can see $\kappa$ correlations on Bamboogle and HotPotQA multi-hop question answering. 
 \begin{table}[ht]
\resizebox{0.48\textwidth}{!}{
\begin{tabular}{lccccc}
\toprule
 & EM & GPT-3.5 & Cmd-R & Haiku & \acronym \\
\midrule
Haiku & 0.746 & 0.772 & 0.741 & 0.802 & \textbf{0.803} \\
Sonnet & 0.767 & 0.879 & 0.880 & 0.877 & \textbf{0.896} \\
CMD-R & 0.599 & 0.881 & 0.896 & 0.897 & \textbf{0.938} \\
Mistral-LG & 0.841 & 0.760 & \textbf{0.936} & \textbf{0.936} & 0.872 \\
GPT-3.5 & 0.849 & 0.886 & \textbf{0.935} & \textbf{0.935} & \textbf{0.935 }\\
GPT-4 & 0.677 & 0.862 & 0.877 & 0.877 & \textbf{0.903} \\ \midrule
Total & 0.762 & 0.846 & 0.885 & \textbf{0.898} & 0.896 \\ \bottomrule
\end{tabular}
}
\caption{Kappa values for various chatbot models on Bamboogle\label{tab:kappa-mh-bamboogle}}
\end{table}

\begin{table}[ht]
\resizebox{0.48\textwidth}{!}{
\begin{tabular}{lccccc}
 & EM & GPT-3.5 & Cmd-R & Haiku & \acronym \\
\toprule
Haiku & 0.703 & \textbf{0.910} & 0.890 & 0.864 & 0.889 \\
Sonnet & 0.715 & 0.796 &\textbf{ 0.847} & 0.844 & 0.820 \\
CMD-R & 0.780 & 0.815 & \textbf{0.909} & 0.906 & 0.859 \\
Mistral-LG & 0.788 & 0.754 &\textbf{ 0.895} & 0.894 & 0.853 \\
GPT-3.5 & 0.799 & 0.757 &\textbf{ 0.879} &\textbf{ 0.879} & 0.858 \\
GPT-4 & 0.768 & \textbf{0.890 }& 0.835 & 0.835 & \textbf{0.890 }\\ \midrule
Total & 0.773 & 0.827 &	\textbf{0.886} & 0.849 & 0.871 \\ \bottomrule
\end{tabular}
}
\caption{Kappa values for various chatbot models on multihop HotPotQA\label{tab:kappa-mh-hpqa}}
\end{table}

\subsubsection{Arena Hard Scores}
Raw scores for Arena-hard coming from \acronym are shown in Table \ref{tab:arena_hard_scores}.
\begin{table}[ht]
\centering
\begin{tabular}{lcc}
\toprule
\textbf{Model} & \textbf{Score} & \textbf{95\% CI} \\ \midrule
GPT-4-turbo & 68.7 & (-2.1, 2.2) \\
Sonnet & 57.6 & (-3.2, 2.2) \\
CMD-R+ & 57.1 & (-2.9, 2.5) \\
Haiku & 55.9 & (-2.1, 2.2) \\
GPT-4-0314 & 50.0 & (0.0, 0.0) \\
Mistral-MD & 49.0 & (-2.3, 2.7) \\
CMD-R & 48.0 & (-1.9, 2.7) \\
Mistral-LG & 43.7 & (-1.9, 2.2) \\
GPT-3.5 & 41.2 & (-2.1, 2.5) \\
\bottomrule
\end{tabular}
\caption{Scores from Arena Hard as scored by the \acronym using average pooling. \label{tab:arena_hard_scores}}
\end{table}

\subsection{Human Annotations \label{sec:appendix_annotation}}
Annotations were performed by profession annotators with diverse demographic and professional backgrounds including novelists, copywriters, copy editors, and journalists. Single-hop TriviaQA, Bamboogle, and multi-hop HotPotQA were singly annotated while Natural Questions and single-hop HotPotQA are triple annotated. Experiments on triple annotated datasets were based on a pooled super-annotator decision based on majority voting. The total number of annotations is shown in Table \ref{tab:num_annotations}.

\begin{table*}[ht]
\centering
\caption{Total Number of Human Annotations by Dataset \label{tab:num_annotations}}
\begin{tabular}{lcc}
\toprule
\textbf{Dataset} & \textbf{Annotated Examples} & \textbf{Annotations per Example}\\ \midrule
Natural Questions & 688 & Triple \\
TriviaQA & 316  & Single\\
HotPotQA (single) & 784 & Triple \\
HotPotQA (multi) & 595  & Single\\
Bamboogle & 750 & Single\\ \bottomrule
\end{tabular}
\end{table*}

\begin{table*}[ht]
    \centering
    \small
    \begin{tabular}{>{\raggedright\arraybackslash\tt}p{\textwidth}<{}}
        \toprule
        Write a high-quality answer for the given question using only the provided search results (some of which might be irrelevant).
\\
\\
Document [1](Title: Lake Eyre basin) keeping pace with evaporation. In contrast, the flow of the Mississippi could fill Lake Eyre in 22 days, that of the Amazon in just 3 days.
Other lakes in the basin include Lake Frome, Lake Yamma Yamma and Lake Hart.
Geography Rivers.
The Cooper Creek, Finke River, Georgina River and Diamantina River are the four main rivers of the basin. Other desert rivers include the Hale River, Plenty River and Todd River that flow from the south east of the Northern Territory, south. In the\\
Document [2](Title: Lake Eyre basin) make it as far south as Lake Eyre, although the story is told that this happened once early in the 20th century. In extreme events, water from the Finke River flows into the Macumba River, which empties into Lake Eyre, a total distance from headwater streams of around . Major tributaries include Ellery Creek, and the Palmer and Hugh Rivers.
The Georgina River system originates on the Barkly Tableland, near the Northern Territory-Queensland border, north-west of Mount Isa and not far south of the Gulf of\\
\\
...
\\
\\
Document [10](Title: Lake Eyre) from the north-east part of the Lake Eyre Basin—in outback (south-west and central) Queensland—flow towards the lake through the Channel Country. The amount of water from the monsoon determines whether water will reach the lake and, if it does, how deep the lake will get. The average rainfall in the area of the lake is per year.
The altitude usually attributed to Kati Thanda–Lake Eyre refers to the deepest parts of the lake floor, in Belt Bay and the Madigan Gulf
\\
\\
Question: where does the water come from to fill lake eyre
\\Answer:
\\
        \bottomrule
    \end{tabular}
    \caption{
        Example of a single-hop question answering prompt from~\citet{liu2024lost} used for KILT answer generations where LLM API does not have a documents parameter.
    }
    \label{tab:litm_prompt}
\end{table*}

\begin{table*}[ht]
    \centering
\begin{adjustbox}{max width=\textwidth, center}
\tiny
    \begin{tabular}{>{\raggedright\arraybackslash\tt}p{1\textwidth}<{}}
        \toprule
You will be given a Question and a Provided Answer. Judge whether the Provided Answer is correct by comparing it to the Reference Answer. Differently formatted dates, people with missing middle names, and alternative spellings should all be considered the same. If the Provided Answer is correct say exactly "True", otherwise say "False".\\
\\
Question 1: "When did the president who set the precedent of a two term limit leave office?"\\
Provided Answer: "George Washington set the precedent of a two-term limit when he decided not to seek a third term in 1796. He left office in 4 March, 1797."\\
Reference Answer: "March 4, 1797"\\
Correct: True\\
\\
\\
Question 2: "Where does Śivarāma Swami conduct courses on Vaishnava Theology?"\\
Provided Answer: "Śivarāma Swami conducts courses on Vaishnava Theology at Bhaktivedanta Manor."\\
Reference Answer: "Where does Śivarāma Swami conduct courses on Vaishnava Theology?"\\
Correct: False\\
\\
\\
Question 3: "The most populous city in Punjab is how large (area wise)?"\\
Provided Answer: "Ludhiana, the most populous city in Punjab, covers an area of 3,767 square kilometres according to the website of Punjab's Department of Agriculture \& Farmer Welfare. Another source states it is 310 square kilometres, making it the most densely populated urban centre in the state."\\
Reference Answer: "310 square kilometers"\\
Correct: True\\
\\
\\
Question 4: "Who was mayor of New York City when Fiorello H. La Guardia was born?"
Provided Answer: "Frank Edson was mayor of New York City when Fiorello H. La Guardia was born on December 11, 1882. Edson served as mayor from 1881 to 1882, and La Guardia himself later became the 99th Mayor of New York City, serving from 1934 to 1946 or 1945."\\
Reference Answer: "William Russell Grace"\\
Correct: False\\
\\
\\
Question 5: "What is the third letter of the top level domain of the military?"\\
Provided Answer: "The third letter of the top-level domain for the military, '.mil', is 'l'. 

The domain name 'mil' is derived from the word 'military'. It is the only top-level domain for a country's military, a legacy of the US military's role in the creation of the internet."\\
Reference Answer: "l"\\
Correct: True\\
\\
\\
Question 6: "Ed Oxenbould plays Tyler in a film directed by which famous director?"\\
Provided Answer: "Ed Oxenbould played Tyler in *The Visit*"\\
Reference Answer: "M. Night Shyamalan"\\
Correct: False\\
\\
\\
Question 7: "Alice David is the voice of Lara Croft in a video game developed by which company ?"\\
Provided Answer: "Alice David is the voice of Lara Croft in the French dub of the video game *Tomb Raider*."\\
Reference Answer: "Crystal Dynamics"\\
Correct: False\\
\\
\\
Question 8: "\{QUESTION\}"\\
Provided Answer: "\{GEN\_ANSWER\}"\\
Reference Answer: "\{GOLD\_ANSWER\}"\\
Correct:\\
        \bottomrule
    \end{tabular}
\end{adjustbox}
    \caption{Multihop Judge Prompt}
    \label{tab:multihop_judge_prompt}
\end{table*}

\begin{table*}[ht]
    \centering
\begin{adjustbox}{max width=\textwidth, center}
\tiny
    \begin{tabular}{>{\raggedright\arraybackslash\tt}p{1.3\textwidth}<{}}
        \toprule
You will be given a Question and a Provided Answer. Judge whether the Provided Answer is correct by comparing it to the Reference Answer. Differently formatted dates, people with missing middle names, and alternative spellings should all be considered the same. If the Provided Answer is correct say exactly "True", otherwise say "False".\\
\\
Question 1: "when did the nfl start playing in london"\\
Provided Answer: "According to the provided search results, the NFL started playing regular season games in London as part of the NFL International Series in 2007. Specifically:\\
\\
Document [5] states: "The NFL International Series was inaugurated in 2007 to host NFL regular season games outside the United States. Played at the new Wembley Stadium in London (rebuilt and reopened in 2007), the series increased from one to two games for the 2013 season, to three games for the 2014 season, and then to four games from the 2017 season."\\
\\
Document [9] also mentions: "Since 2007, the league has held multiple regular season games in London each season as part of NFL London Games, allowing the league to test solutions to some of the challenges facing a hypothetical London franchise."\\
\\
So 2007 marked the start of the NFL playing regular season games at the newly rebuilt Wembley Stadium in London as part of the International Series. Prior to that, there were some preseason games played in London, but 2007 was the first year for regular season NFL games in the city."\\
Reference Answer: "since 2007"\\
Correct: True\\
\\
\\
Question 2: "where is wind power used most in the united states"\\
Provided Answer: "Iowa is the state that generates the highest percentage of energy from wind power, with more than 35\% of its electric power coming from wind. Alternatively, North Dakota has the highest per capita wind generation. \\
\\
The state with the most installed wind power capacity is Texas, followed by Iowa. Texas's capacity is around 20,000 MW, almost double that of Iowa's 3,670 MW. The Roscoe Wind Farm in Texas is also the world's largest wind farm at 780 MW. \\
\\
California has doubled its wind power capacity since 2002 and now gets about 5\% of its energy from wind power. This is still lower than the twelve U.S. states that get more than 20\% of their energy from wind."\\
Reference Answer: "Texas"\\
Correct: True\\
\\
\\
Question 3: "who is the owner of phoenix mall in chennai"\\
Provided Answer: "The owner of Phoenix Marketcity in Chennai is Phoenix Mills Limited, as mentioned in the document titled "Phoenix Marketcity (Chennai)." This mall was jointly developed by Phoenix Mills Limited and Crest Ventures Ltd."\\
Reference Answer: "Phoenix Mills Limited"\\
Correct: True\\
\\
\\
Question 4: "number of degree of freedom for plane mechanism"\\
Provided Answer: "In mechanics, the degree of freedom (DOF) of a mechanical system is the number of independent parameters that define its configuration. The number of degrees of freedom for a mechanism on a two-dimensional plane is three, which consists of two translations and one rotation. \\
\\
For a system of multiple bodies, the combined DOF is the sum of the DOFs of the individual bodies, minus the internal constraints on their relative motion. Therefore, a mechanism formed of multiple rigid bodies may have more than three DOFs."\\
Reference Answer: "two"\\
Correct: False\\
\\
\\
Question 5: "who had the best strike rate among batsmen who scored over 200 runs in the world cup"\\
Provided Answer: "The provided documents do not specify the strike rate of all batsmen who scored over 200 runs in the World Cup. However, among the batsmen mentioned who have scored over 200 runs in the World Cup, Jehan Mubarak has the highest ever strike rate in a T20 World Cup match for Sri Lanka, with a strike rate of 353.84 (Document [6]). But it's important to note that this strike rate is not necessarily reflective of his overall performance in the World Cup, as it could have been achieved in a single match.

For a more accurate comparison of strike rates among batsmen who scored over 200 runs in the World Cup, a comprehensive list of all such batsmen and their respective strike rates would be required."\\
Reference Answer: "Rohit Sharma"\\
Correct: False\\
\\
\\
Question 6: "who played ben stone son on law and order"\\
Provided Answer: "According to the provided search results, Ben Stone's son Peter Stone was played by Philip Winchester. Some key details:

- Peter Stone is the son of Benjamin "Ben" Stone, the former Executive Assistant District Attorney portrayed by Michael Moriarty on Law \& Order.

- Peter first appeared as a character on the Chicago P.D. episode "Justice", which served as a backdoor pilot for the short-lived spin-off Chicago Justice, where he was the lead character played by Philip Winchester. 

- After Chicago Justice was canceled, Philip Winchester reprised his role as Peter Stone in Season 19 of Law \& Order: Special Victims Unit, where he became the Assistant District Attorney for the Sex Crimes Bureau.

- In several SVU episodes, Peter describes having a strained relationship with his workaholic father Ben Stone, who neglected his family in favor of his prosecutorial duties.

So in summary, Philip Winchester played the role of Peter Stone, the son of the iconic Law \& Order character Ben Stone, across the shows Chicago Justice and Law \& Order: SVU."\\
Reference Answer: "Michael Moriarty"\\
Correct: False\\
\\
\\
Question 7: "who won so you think you can dance season 13"\\
Provided Answer: "I'm sorry, I couldn't find any information about the winner of season 13 of So You Think You Can Dance."
Reference Answer: "Leon `` Kida '' Burns"\\
Correct: False\\
\\
\\
Question 8: "when did sweet caroline start at red sox games"\\
Provided Answer: "The song "Sweet Caroline" started being played at Fenway Park, home of the Boston Red Sox, since at least 1997. However, it has been played consistently in the middle of the eighth inning at every game since 2002."\\
Reference Answer: "at least 1997"\\
Correct: True\\
\\
\\
Question 9: "when did end of the road come out"\\
Provided Answer: "I found multiple results for "End of the Road":
\\
\#\# End of the Road (1944 film)\\
End of the Road is a 1944 American crime film directed by George Blair and written by Denison Clift, Gertrude Walker and Albert Beich. The film stars Edward Norris, John Abbott, June Storey, Jonathan Hale, Pierre Watkin and Ted Hecht.
\\
\#\# End of the Road (1970 film)\\
End of the Road is a 1970 American comedy-drama film directed, co-written, and edited by Aram Avakian and adapted from a 1958 novel by John Barth. The film stars Stacy Keach, James Earl Jones and Harris Yulin.
\\
\#\# End of the Road (Boyz II Men song)\\
End of the Road is a single recorded by American R\&B group Boyz II Men for the "Boomerang" soundtrack. It was released on June 30, 1992.\\
\\
\#\# End of the Road (Jerry Lee Lewis song)\\
End of the Road is a song by Jerry Lee Lewis. It appeared on the 1976 Sun Records compilation album "The Original Jerry Lee Lewis" which was released by the Sun Record Company.
\\
Which of these were you referring to?"\\
Reference Answer: "June 30 , 1992"\\
Correct: True\\
\\
\\
Question 10: "who played john connor in the original terminator"\\
Provided Answer: "John Connor is not physically portrayed by any actor in the original "Terminator" film released in 1984, as he is only mentioned and not seen. The character of John Connor is first referred to in this film as the future leader of the human resistance against the machines, but he does not make an appearance."\\
Reference Answer: "Edward Furlong throughout the remainder of the film"\\
Correct: False\\
\\
\\
Question 11: "\{QUESTION\}"\\
Provided Answer: "\{GEN\_ANSWER\}"\\
Reference Answer: "\{GOLD\_ANSWER\}"\\
Correct:\\
        \bottomrule
    \end{tabular}
\end{adjustbox}
    \caption{KILT NaturalQuestions judge prompt.
    }
    \label{tab:nq_kilt_prompt}
\end{table*}

\begin{table*}[ht]
    \centering
\begin{adjustbox}{max width=\textwidth, center}
\tiny
    \begin{tabular}{>{\raggedright\arraybackslash\tt}p{1.5\textwidth}<{}}
        \toprule
        You will be given a Question and a Provided Answer. Judge whether the Provided Answer is correct by comparing it to the Reference Answer. Differently formatted dates, people with missing middle names, and alternative spellings should all be considered the same. If the Provided Answer is correct say exactly "True", otherwise say "False".\\
\\
Question 1: "Who was also an actor, Serri or John Fogerty?"\\
Provided Answer: "Based on the provided search results, there is no information indicating that John Fogerty has pursued acting as a profession or has been involved in acting projects. The search results focus on his music career, including his time with Creedence Clearwater Revival, his solo career, and various performances and recognitions he has received over the years. Therefore, without information on Serri in the provided documents, it cannot be determined if Serri was an actor. However, it is clear from the search results that John Fogerty is primarily known as a musician, singer, and songwriter, not as an actor."\\
Reference Answer: "Serri"\\
Correct: False\\
\\
\\
Question 2: "The 37th Disney animated feature film Tarzan features You'll be in my heart by which artist?"
Provided Answer: "The 37th Disney animated feature film Tarzan's soundtrack, Tarzan: An Original Walt Disney Records, includes the song You'll Be in My Heart written and performed by Phil Collins. The song won the Academy Award for Best Original Song and the Golden Globe Award for Best Original Song. Collins was originally hired as a songwriter but ended up performing the song as well."\\
Reference Answer: "Phil Collins"\\
Correct: True\\
\\
\\
Question 3: "Who is the current drummer of the band who did the song "What Lovers Do"?"\\
Provided Answer: "Unfortunately, none of the provided search results directly mention the current drummer for the band Maroon 5 who performed the song "What Lovers Do". The results provide background information on the song itself, its music video, chart performance, and other details, but do not specify the band members or drummer.\\
\\
Maroon 5 is a well-known pop rock band fronted by Adam Levine, but their full current lineup and specific drummer for the "What Lovers Do" era is not stated in these search results. To definitively answer who the drummer was, additional information directly about Maroon 5\'s band members around 2017 when "What Lovers Do" was released would be needed."
Reference Answer: "Matt Flynn"\\
Correct: False\\
\\
\\
Question 4: "Who is older, Ferdi Taygan or Mahesh Bhupathi?"\\
Provided Answer: "Unfortunately, none of the provided search results contain any information about Ferdi Taygan or their age to compare with Mahesh Bhupathi's age. The search results are all focused on providing biographical details about Mahesh Bhupathi, an Indian professional tennis player, and some other individuals named Mahesh Bhatt who appear to be unrelated to the question. Without any information about Ferdi Taygan, it is impossible to determine who is older between them and Mahesh Bhupathi based solely on these search results."\\
Reference Answer: "Ferdi Taygan"\\
Correct: False\\
\\
\\
Question 5: "What is the birthday of the actress who was the Duchess in "The Revengers Tragedy"?"\\
Provided Answer: "Diana Quick, who played the Duchess in "The Revengers Tragedy", was born on 23rd September 1934."\\
Reference Answer: "23 November 1946"\\
Correct: False\\
\\
\\
Question 6: "Rejuvelac is kind of grain water invented and promoted by a "holistic health" practitioner born in which year ?"\\
Provided Answer: "Rejuvelac is a kind of grain water invented and promoted by Ann Wigmore, who was born in 1909."\\
Reference Answer: "1909"\\
Correct: True\\
\\
\\
Question 7: "Who is the current drummer of the band who did the song "What Lovers Do"?"\\
Provided Answer: "The song "What Lovers Do" is by the American pop rock band Maroon 5. The current drummer of Maroon 5 is Matt Flynn."\\
Reference Answer: "Matt Flynn"\\
Correct: True\\
\\
\\
Question 8: "The 37th Disney animated feature film Tarzan features You'll be in my heart by which artist?"\\
Provided Answer: "The song "You\'ll Be in My Heart" featured in the 1999 Disney animated film Tarzan was written and performed by Phil Collins.

The key details that support this answer from the provided search results:

1. Document [2] states: "You\'ll Be in My Heart" is a song by Phil Collins, from the 1999 Disney animated feature "Tarzan"."\\
\\
2. Document [3] mentions: "For his contribution to the soundtrack, Collins received an American Music Award for Favorite Adult Contemporary Artist."\\
\\
3. Document [5] says: "Collins was originally hired by Disney as a songwriter, but was chosen because he was a drummer at first and given his history as a drummer for the rock band Genesis, he appealed to the Disney production crew, who "wanted a strong jungle beat to accompany Tarzan\'s adventures". This "ballad" is one of five original songs he wrote for the film, and he ended up also performing this song as well."\\
\\
4. Document [7] states: "The song is about "how love is a bond that cannot be broken"."\\
\\
So in summary, the artist who wrote and performed the song "You\'ll Be in My Heart" featured in the Disney animated film Tarzan is Phil Collins."\\
Reference Answer: "Phil Collins"\\
Correct: True\\
\\
\\
Question 9: "How long is the river for which Frenchmans Creek is a tributary?"\\
Provided Answer: "The length of the river for which Frenchmans Creek is a tributary cannot be directly determined from the provided documents as they only provide information about the Frenchmans Creek itself and not the river it flows into. However, we can see that Frenchmans Creek is a tributary of the Darling River in New South Wales, Australia (Document 2), the Mississippi River in Missouri, USA (Document 3), and the Milk River in Saskatchewan, Canada, and Montana, USA (Document 5). The length of these rivers is not provided in the documents."\\
Reference Answer: "2844 km"\\
Correct: False\\
\\
\\
Question 10: "Conrad Anker located the body of a mountaineer who took part it the first three British expeditions to which mountain?"\\
Provided Answer: "Conrad Anker located the body of George Mallory on Mount Everest. Mallory was a British climber who took part in the first three British expeditions to Mount Everest, including the 1924 expedition during which he and his climbing partner Andrew Irvine disappeared while attempting to reach the summit. Anker found Mallory's body in 1999 as a member of a search team looking for the remains of the British climber.

References:
[1] Conrad Anker
[2] 2007 Altitude Everest expedition
[9] Timeline of Mount Everest expeditions"\\
Reference Answer: "Mount Everest"\\
Correct: True\\
\\
\\
Question 11: "The city that contains the Yunnan Provincial Museum is also known by what nickname?"\\
Provided Answer: "The city that contains the Yunnan Provincial Museum, Kunming, is also known by the nickname "Spring City.""\\
Reference Answer: "Spring city"\\
Correct: True\\
\\
\\
Question 12: "The 1st Virginia Regiment saw action at a battle that led to the departure of what Royal Governor?"
Provided Answer: "Based on the provided search results, the 1st Virginia Regiment saw action at the Battle of Great Bridge in December 1775, which led to the departure of Lord Dunmore, the Royal Governor of Virginia at the time.\\
\\
Document [1] mentions that the 1st Virginia Regiment participated in the Battle of Great Bridge. Document [4] states that after the Battle of Great Bridge, "Dunmore responded by bombarding Norfolk with his ships on January 1, 1776." This implies that the Battle of Great Bridge, in which the 1st Virginia Regiment fought, led to Lord Dunmore\'s departure from Virginia as the Royal Governor.\\
\\
So the 1st Virginia Regiment\'s involvement in the Battle of Great Bridge resulted in the departure of Lord Dunmore, the Royal Governor of the Colony of Virginia during the American Revolutionary War."\\
Reference Answer: "Royal Governor Lord Dunmore"\\
Correct: True\\

Question 13: "The voice of Homer Simpson also voices what alien character in the television series "The Simpsons"?"\\
Provided Answer: "Dan Castellaneta voices Homer Simpson, Grampa Simpson, Krusty the Clown, Groundskeeper Willie, Mayor Quimby, Barney Gumble, and other adult characters."\\
Reference Answer: "Kodos"\\
Correct: False\\
\\
\\
Question 14: "Who were the producers of the show for which Chris McKay made his feature film directorial debut?"
Provided Answer: "Chris McKay made his feature film directorial debut with "The Lego Batman Movie" in 2017. The producers of "The Lego Batman Movie" were not explicitly mentioned in the provided documents. However, it is noted that Seth Grahame-Smith co-wrote the screenplay for the movie, and Warner Bros. was involved in its development, as mentioned in Document 8. Additionally, Document 9 mentions that McKay had signed a first look deal for being a producer with Warner Bros. around the time of the movie\'s development. While this information does not directly answer the question regarding the specific producers of "The Lego Batman Movie," it does highlight the key individuals and entities involved in its creation and development."
Reference Answer: "Dan Lin, Roy Lee, Phil Lord and Christopher Miller"
Correct: False\\
\\
\\
Question 15: "Marion Greene was a health policy analyst for St. Judt Medical company, which had how many principal operations worldwide?"\\
Provided Answer: "The provided documents do not specify the number of principal operations St. Jude Medical company had worldwide. Document 1 mentions that Marion Greene was a health policy analyst for St. Jude Medical, but it does not provide details about the company's operations. Document 2 also mentions Greene's work for St. Jude Medical, but again, no information is given about the company's operations. The other documents do not pertain to St. Jude Medical. Therefore, I cannot provide an answer to this question based on the given documents."\\
Reference Answer: "20"\\
Correct: False\\

Question 16: "What is the name of the actress who has received a Drama Desk award and was part of the cast album for the 2007 Broadway production of the hit musical, "Grease"?"\\
Provided Answer: "Laura Osnes, who played the role of Sandy Dumbrowski in the 2007 Broadway production of "Grease", is a Drama Desk award winner and is featured on the cast album for the production."\\
Reference Answer: "Laura Ann Osnes"\\
Correct: True\\
\\
\\
Question 17: "\{QUESTION\}"\\
Provided Answer: "\{GEN\_ANSWER\}"\\
Reference Answer: "\{GOLD\_ANSWER\}"\\
Correct:\\
        \bottomrule
    \end{tabular}
    \end{adjustbox}
    \caption{KILT HotpotQA Judge prompt}
    \label{tab:hotpotqa_kilt_prompt}
\end{table*}

\begin{table*}[ht]
    \centering
\begin{adjustbox}{max width=\textwidth, center}
\tiny
    \begin{tabular}{>{\raggedright\arraybackslash\tt}p{1.2\textwidth}<{}}
        \toprule
    You will be given a Question and a Provided Answer. Judge whether the Provided Answer is correct by comparing it to the Reference Answer. Differently formatted dates, people with missing middle names, and alternative spellings should all be considered the same. If the Provided Answer is correct say exactly "True", otherwise say "False".\\
\\
Question 1: "If you landed at 'Santa Catarina Airport on which Island would you be in?"\\
Provided Answer: "If you landed at Santa Catarina Airport, you would be on the island of Santa Catarina, in the Brazilian state of the same name. The island is approximately 54km long and 18km wide and is home to the state capital, Florianópolis. It is the largest island in an archipelago of over 30 islands."\\
Reference Answer: "Geography of the Madeira Islands"\\
Correct: False\\
\\
\\
Question 2: "When it was a first performed at the Abbey Theatre in Dublin in 1907, which play caused rioting in the streets because of objections to the portrayal of rural Irish Catholic immorality?"\\
Provided Answer: "Based on the provided search results, the play that caused rioting in the streets when it was first performed at the Abbey Theatre in Dublin in 1907 due to objections over the portrayal of rural Irish Catholic immorality was "The Playboy of the Western World" by J.M. Synge.\\
\\
Several documents mention the riots that erupted during and after the opening performance of Synge\'s "The Playboy of the Western World" at the Abbey Theatre in January 1907. The riots were instigated by Irish nationalists and republicans who viewed the play\'s content as offensive to public morals and an insult to Ireland (Documents 1, 3, 7). \\
\\
Document 5 specifically states that "the first performance of J.M. Synge\'s play \'The Playboy of the Western World\' at the Abbey Theatre in Dublin triggers a week of rioting" in 1907. Document 9 also notes that a performance of the play on January 28, 1907 "was interrupted by the audience who continue to boo, hiss and shout."\\
\\
The objections seem to have stemmed from the play\'s depiction of patricide and perceived immorality in rural Irish Catholic society, which nationalists found offensive (Document 7). So "The Playboy of the Western World" by J.M. Synge was the controversial play that provoked riots when first staged at the Abbey Theatre in 1907 due to its portrayal of the Irish countryside."\\
Reference Answer: "'THE PLAYBOY OF THE WESTERN WORLD'"\\
Correct: True\\
\\
\\
Question 3: "The religious order of Poor Ladies of San Damiano has what more common name?"
Provided Answer: "The religious order of Poor Ladies of San Damiano is more commonly known as the Poor Clares. This order was founded by Saints Clare of Assisi and Francis of Assisi in 1212, and it was the second Franciscan Order to be established. The Poor Clares were initially referred to as the Order of Poor Ladies, and later the Clarisses, the Minoresses, the Franciscan Clarist Order, and the Second Order of Saint Francis. However, the name Poor Clares is the most recognized and widely used today."\\
Reference Answer: "Franciscan Clarist Order"\\
Correct: False\\
\\
\\
Question 4: "Which country defeated the Ivory Coast to win football's Africa Cup of Nations?"\\
Provided Answer: "Based on the provided search results, Ghana defeated the Ivory Coast in the final of the 1992 Africa Cup of Nations. The key evidence is from Document 1 and Document 10:\\
\\
Document 1 states: "Ivory Coast set a record for the competition by holding each of their opponents scoreless in the six matches of the final tournament after his contributions helped Ghana reach the final; he was, however, suspended for that match and Ghana lost to Ivory Coast in a penalty shootout that saw each side make 11 attempts to determine the winner."\\
\\
Document 10 directly mentions: "The 1992 Africa Cup of Nations Final was a football match that took place on January 26, 1992, at the Stade de l\'Amitié in Dakar, Senegal, to determine the winner of the 1992 African Cup of Nations. Ivory Coast defeated Ghana 11–10 on penalties after a goalless draw to win their first African Cup."\\
\\
So the Ivory Coast defeated Ghana in the 1992 Africa Cup of Nations final after a penalty shootout. None of the other provided documents contradict this or indicate another country defeated the Ivory Coast to win that specific tournament."\\
Reference Answer: "health care in zambia"\\
Correct: False\\
\\
\\
Question 5: "Fred Scuttle was one of the best-known characters of which British comedian?"\\
Provided Answer: "The provided documents do not directly mention Fred Scuttle or the British comedian associated with this character. Therefore, based on the information available in the documents, it is not possible to accurately answer the question about which British comedian is best known for the character Fred Scuttle."\\
Reference Answer: "Alfred Hawthorne Hill"\\
Correct: False\\
\\
\\
Question 6: "What element is the main constituent of anthracite?"\\
Provided Answer: "The main constituent of anthracite is carbon. High-grade anthracite, for example, has a chemical formula represented by CHONS, indicating it contains carbon, hydrogen, oxygen, nitrogen, and sulfur, with carbon making up 94\% of its composition. Ultra high-grade (UHG) anthracite typically has a minimum carbon content of 95\%, further emphasizing carbon as the primary element in anthracite."\\
Reference Answer: "Carbon"\\
Correct: True\\
\\
\\
Question 7: "Which American played in his 10th Ryder Cup in 2014?"\\
Provided Answer: "Mickelson extended his own U.S. team record with his 10th appearance at the 2014 Ryder Cup."\\
Reference Answer: "philip alfred mickelson"\\
Correct: True\\
\\
\\
Question 8: "Who is the lead singer of the Sheffield band the Arctic Monkeys?"\\
Provided Answer: "The lead singer of the Sheffield band Arctic Monkeys is Alex Turner. Turner also plays guitar and is the frontman of the four-piece indie rock band.\\
Arctic Monkeys have released six studio albums, including a debut in 2002 titled "Whatever People Say I Am, That's What I'm Not"."\\
Reference Answer: "Alex Turner"\\
Correct: True\\
\\
\\
Question 9: "\{QUESTION\}"\\
Provided Answer: "\{GEN\_ANSWER\}"\\
Reference Answer: "\{GOLD\_ANSWER\}"\\
Correct:\\
        \bottomrule
    \end{tabular}
    \end{adjustbox}

    \caption{
    KILT TriviaQA Judge Prompt
    }
    \label{tab:triviaqa_kilt_prompt}
\end{table*}

\begin{table*}[ht]
    \centering
\begin{adjustbox}{max width=\textwidth, center}
\tiny
    \begin{tabular}{>{\raggedright\arraybackslash\tt}p{1.3\textwidth}<{}}
        \toprule
SYSTEM: You are judging whether a model has generated a correct answer to a question. Study the examples the user gives you as they will be very informative for how to do the task. The Reference Answers you get will be short. An model's answer will be longer, and can be considered correct if it contains the semantic content of short reference answer somewhere within it. Don't worry about factuality with respect to the real world, just judge the example based on what you see. No need to overthink this task, it really comes down to just soft matching. Answer with only the word 'Yes' or 'No'\\
\\
USER: Question: when did the nfl start playing in london
Model Answer: According to the provided search results, the NFL started playing regular season games in London as part of the NFL International Series in 2007. Specifically:\\
\\
Document [5] states: "The NFL International Series was inaugurated in 2007 to host NFL regular season games outside the United States. Played at the new Wembley Stadium in London (rebuilt and reopened in 2007), the series increased from one to two games for the 2013 season, to three games for the 2014 season, and then to four games from the 2017 season."\\
\\
Document [9] also mentions: "Since 2007, the league has held multiple regular season games in London each season as part of NFL London Games, allowing the league to test solutions to some of the challenges facing a hypothetical London franchise."\\
\\
So 2007 marked the start of the NFL playing regular season games at the newly rebuilt Wembley Stadium in London as part of the International Series. Prior to that, there were some preseason games played in London, but 2007 was the first year for regular season NFL games in the city.\\
Reference Answer: since 2007\\
Correct: Yes\\
-----\\
Question: where is wind power used most in the united states\\
Model Answer: Iowa is the state that generates the highest percentage of energy from wind power, with more than 35\% of its electric power coming from wind. Alternatively, North Dakota has the highest per capita wind generation. \\
\\
The state with the most installed wind power capacity is Texas, followed by Iowa. Texas's capacity is around 20,000 MW, almost double that of Iowa's 3,670 MW. The Roscoe Wind Farm in Texas is also the world's largest wind farm at 780 MW. \\
\\
California has doubled its wind power capacity since 2002 and now gets about 5\% of its energy from wind power. This is still lower than the twelve U.S. states that get more than 20\% of their energy from wind.\\
Reference Answer: Texas\\
Correct: Yes\\
-----\\
Question: who is the owner of phoenix mall in chennai\\
Model Answer: The owner of Phoenix Marketcity in Chennai is Phoenix Mills Limited, as mentioned in the document titled "Phoenix Marketcity (Chennai)." This mall was jointly developed by Phoenix Mills Limited and Crest Ventures Ltd.\\
Reference Answer: Phoenix Mills Limited\\
Correct: Yes\\
-----\\
Question: number of degree of freedom for plane mechanism\\
Model Answer: In mechanics, the degree of freedom (DOF) of a mechanical system is the number of independent parameters that define its configuration. The number of degrees of freedom for a mechanism on a two-dimensional plane is three, which consists of two translations and one rotation. 
\\
For a system of multiple bodies, the combined DOF is the sum of the DOFs of the individual bodies, minus the internal constraints on their relative motion. Therefore, a mechanism formed of multiple rigid bodies may have more than three DOFs.\\
Reference Answer: two\\
Correct: No\\
-----\\
Question: who had the best strike rate among batsmen who scored over 200 runs in the world cup\\
Model Answer: The provided documents do not specify the strike rate of all batsmen who scored over 200 runs in the World Cup. However, among the batsmen mentioned who have scored over 200 runs in the World Cup, Jehan Mubarak has the highest ever strike rate in a T20 World Cup match for Sri Lanka, with a strike rate of 353.84 (Document [6]). But it's important to note that this strike rate is not necessarily reflective of his overall performance in the World Cup, as it could have been achieved in a single match.\\
\\
For a more accurate comparison of strike rates among batsmen who scored over 200 runs in the World Cup, a comprehensive list of all such batsmen and their respective strike rates would be required.\\
Reference Answer: Rohit Sharma\\
Correct: No\\
-----\\
Question: who played ben stone son on law and order\\
Model Answer: According to the provided search results, Ben Stone's son Peter Stone was played by Philip Winchester. Some key details:\\
\\
- Peter Stone is the son of Benjamin "Ben" Stone, the former Executive Assistant District Attorney portrayed by Michael Moriarty on Law \& Order.\\
\\
- Peter first appeared as a character on the Chicago P.D. episode "Justice", which served as a backdoor pilot for the short-lived spin-off Chicago Justice, where he was the lead character played by Philip Winchester. \\
\\
- After Chicago Justice was canceled, Philip Winchester reprised his role as Peter Stone in Season 19 of Law \& Order: Special Victims Unit, where he became the Assistant District Attorney for the Sex Crimes Bureau.\\
\\
- In several SVU episodes, Peter describes having a strained relationship with his workaholic father Ben Stone, who neglected his family in favor of his prosecutorial duties.\\
\\
So in summary, Philip Winchester played the role of Peter Stone, the son of the iconic Law \& Order character Ben Stone, across the shows Chicago Justice and Law \& Order: SVU.\\
Reference Answer: Michael Moriarty\\
Correct: No\\
-----\\
Question: who won so you think you can dance season 13\\
Model Answer: I'm sorry, I couldn't find any information about the winner of season 13 of So You Think You Can Dance.\\
Reference Answer: Leon `` Kida '' Burns\\
Correct: No\\
-----
Question: when did sweet caroline start at red sox games\\
Model Answer: The song "Sweet Caroline" started being played at Fenway Park, home of the Boston Red Sox, since at least 1997. However, it has been played consistently in the middle of the eighth inning at every game since 2002.\\
Reference Answer: at least 1997\\
Correct: Yes\\
-----\\
Question: when did end of the road come out\\
Model Answer: I found multiple results for "End of the Road":\\
\\
\#\# End of the Road (1944 film)\\
End of the Road is a 1944 American crime film directed by George Blair and written by Denison Clift, Gertrude Walker and Albert Beich. The film stars Edward Norris, John Abbott, June Storey, Jonathan Hale, Pierre Watkin and Ted Hecht.\\
\\
\#\# End of the Road (1970 film)\\
End of the Road is a 1970 American comedy-drama film directed, co-written, and edited by Aram Avakian and adapted from a 1958 novel by John Barth. The film stars Stacy Keach, James Earl Jones and Harris Yulin.\\
\\
\#\# End of the Road (Boyz II Men song)\\
End of the Road is a single recorded by American R\&B group Boyz II Men for the "Boomerang" soundtrack. It was released on June 30, 1992.\\
\\
\#\# End of the Road (Jerry Lee Lewis song)\\
End of the Road is a song by Jerry Lee Lewis. It appeared on the 1976 Sun Records compilation album "The Original Jerry Lee Lewis" which was released by the Sun Record Company.\\
\\
Which of these were you referring to?\\
Reference Answer: June 30 , 1992\\
Correct: Yes\\
-----\\
Question: who played john connor in the original terminator\\
Model Answer: John Connor is not physically portrayed by any actor in the original "Terminator" film released in 1984, as he is only mentioned and not seen. The character of John Connor is first referred to in this film as the future leader of the human resistance against the machines, but he does not make an appearance.
Reference Answer: Edward Furlong throughout the remainder of the film\\
Correct: No\\
-----\\
Question: \{QUESTION\}\\
Model Answer: \{GEN\_ANSWER\}\\
Reference Answer: \{GOLD\_ANSWER\}\\
Correct:\\
        \bottomrule
    \end{tabular}
    \end{adjustbox}

    \caption{
    KILT NQ  GPT-4 optimized Judge Prompt
    }
    \label{tab:nq_kilt_prompt_gpt4_opt}
\end{table*}

\end{document}